\documentclass{article}
\usepackage{arxiv}
\usepackage{times}
\usepackage{soul}
\usepackage[utf8]{inputenc}
\usepackage{graphicx}
\usepackage{booktabs}
\usepackage{algorithm}
\usepackage[noend]{algorithmic}

\usepackage{helvet}
\usepackage{courier}
\usepackage[hyphens]{url}

\usepackage{natbib}
\usepackage{booktabs}
\usepackage{multirow}
\usepackage{makecell}
\usepackage{pifont}
\usepackage{amssymb}
\usepackage{mathtools}
\usepackage{tikz}
\usetikzlibrary{bayesnet}
\usetikzlibrary{decorations.pathmorphing} 
\usetikzlibrary{decorations.markings} 
\usetikzlibrary{matrix} 
\usetikzlibrary{arrows} 
\usetikzlibrary{calc} 
\tikzset{strike through/.append style={
    decoration={markings, mark=at position 0.48 with {
    \draw[-, solid, line width=1.2pt] ++ (-2pt,-3.5pt) -- (2pt,3.5pt);},
    mark=at position 0.52 with {
    \draw[-, solid, line width=1.2pt] ++ (-2pt,-3.5pt) -- (2pt,3.5pt);}
  },postaction={decorate}}
}
\newcommand*{\qed}{\null\nobreak\hfill\ensuremath{\blacksquare}}%

\title{Right for the Right Latent Factors: Debiasing Generative Models via Disentanglement}

\author{
\Large \textbf{Xiaoting Shao\textsuperscript{\rm 1}},
\Large \textbf{Karl Stelzner\textsuperscript{\rm 1}, Kristian Kersting\textsuperscript{\rm 1, \rm 2}} \\
\textsuperscript{\rm 1}CS Department, and \textsuperscript{\rm 2}Centre for Cognitive Science, TU Darmstadt, Germany
}

\begin{document}
\maketitle

\begin{abstract}
A key assumption of most statistical machine learning methods is that they have access to independent samples from the distribution of data they encounter at test time. As such, these methods often perform poorly in the face of biased data, which breaks this assumption.
In particular, machine learning models have been shown to exhibit Clever-Hans-like behaviour, meaning that spurious correlations in the training set are inadvertently learnt. A number of works have been proposed to revise deep classifiers to learn the right correlations. However, generative models have been overlooked so far. We observe that generative models are also prone to Clever-Hans-like behaviour. To counteract this issue, we propose to debias generative models by disentangling their internal representations, which is achieved via human feedback. Our experiments show that this is effective at removing bias even when human feedback covers only a small fraction of the desired distribution. In addition, we achieve strong disentanglement results in a quantitative comparison with recent methods.
\end{abstract}

\section{Introduction}
The key assumption behind standard machine learning techniques is that training and test data are 
independently and identically distributed (i.i.d.). In practice, different forms of bias can creep into our datasets
in subtle ways, breaking this assumption \cite{torralba2011unbiased, tommasi2017deeper, ritter2017cognitive, stock2018convnets, grover2019fair}.
As a result, machine learning system can become worthless or even harmful. For instance, a classifier may learn to make
decisions based on a confounder present in the training data and then fail to generalize to the test data
\cite{lapuschkin2019unmasking}. Similarly, a machine learning system may propagate unfair biases present
in the training data by learning to make decisions based on protected attributes such as race or gender
\cite{hardt2016equality}.

A number of methods have been proposed to repair deep classifiers which have learned the wrong rules, or to prevent them from doing so in the first place \cite{ross2017right, murdoch2018beyond, teso2019explanatory, rieger2019interpretations, erion2019learning, selvaraju2019taking, schramowski2020making, shao2021right, kim2019learning, grover2019bias}.
Generative models however have generally not been addressed, despite the fact that they have the potential to amplify bias
by generating more biased data at test time \cite{grover2019fair}.
In this work, we confirm experimentally that generative models are prone to learning biases as well.
A commonly used approach to combat this issue is \emph{data augmentation}. When it is easy to generate synthetic data
in which the confounding variable is randomly varied, then an augmented version of the original training set can be constructed
which no longer exhibits (undesirable) bias \cite{geirhos2018imagenettrained}.
However, when such data needs to be manually collected, this approach is typically prohibitively expensive,
as the number of ways in which factors of variation can be combined grows exponentially \cite{bahng2019rebias}.

\begin{figure}[t]
    \centering
    \includegraphics[width=0.3\columnwidth]{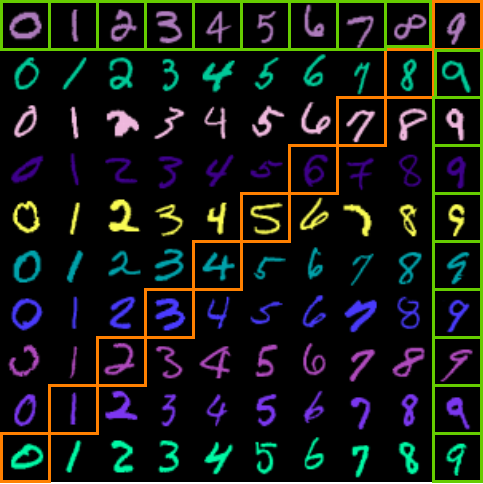} \quad \quad
    \includegraphics[width=0.31\columnwidth, trim={0.5cm, 0.7cm, 0cm, 0cm}, clip]{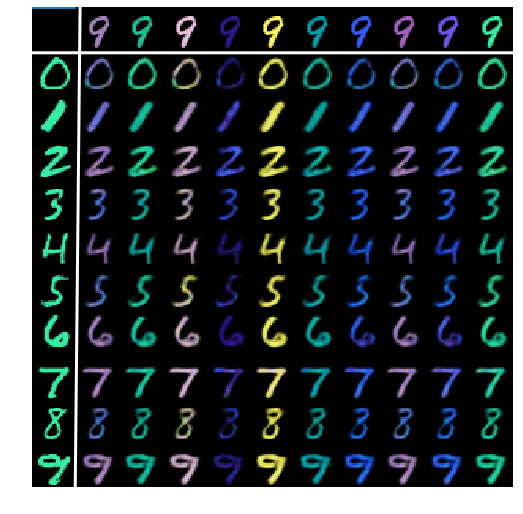}
    \caption{\emph{Left:} Our model is trained on a biased version of the colored MNIST dataset, in which color and shape are fully correlated (diagonal marked in orange). As human feedback, we supply a small number of labelled examples from the first row and the rightmost column.
    \emph{Right:} Despite never observing most combinations of color and shape (observed combinations are marked in orange and green), our model learns to generate all of them. Here, reconstructions are drawn by combining the shape (top row) with the color of the digit (leftmost column).}
    \label{fig:overview}
\end{figure}

In this paper, we show how this issue can be prevented by learning disentangled representations.
A representation is called \emph{disentangled} if its variables correspond directly to the factors of variation underlying the data
\cite{bengio2013representation, van2019disentangled, locatello2019fairness, creager2019flexibly, chartsias2019disentangled, higgins2017darla}.
Specifically, we study variational autoencoders (VAEs) \cite{Kingma2014AutoEncodingVB}. When such a model is presented with
biased data, if no precautions are taken, it will generally match the biased training distribution and learn a representation
in which the underlying factors of variation are entangled.
If we can however disentangle these factors, we can place an independent prior over the 
latent variables and thereby force the models to learn a distribution in which the factors of variation are independent. This is a significant challenge, since in addition to achieving disentanglement, such a model needs to generalize to those combinations of attributes which were not present in the training data due to bias.

To learn such disentangled representation, we use a small amount of human feedback. The feedback technique is built on recent work on weakly supervised disentanglement \cite{chen2019weakly, locatello2020weakly, shu2020weakly}. Specifically, it takes the form of a small amount of labelled data in which one attribute remains constant, while the others are varied. Even though most combinations are never observed by the model, it nevertheless learns to generate, reconstruct, and encode such examples correctly, as illustrated in Fig.~\ref{fig:overview}. We find that generalization can also be achieved with this small amount of feedback. This addresses the often prohibitive cost of data augmentation.

Our key contributions are:
    (1) We propose a method for learning unbiased, disentangled representations from severely biased data by using human feedback within the VAE framework. 
    (2) We formulate human feedback as a set of examples with partially observed labels to disentangle the internal representations of VAEs. 
    (3) We prove that this approach can guarantee restrictiveness and non-trivial solutions when disentangling a subset of factors, accommodating for possible nuisance factors. 
    (4) We demonstrate empirically on several benchmarks that our unbiased generative model generalizes to novel combinations of the underlying factors of variation. The disentangled latent representation allows fine-grained control over the generated samples.

We proceed by touching upon related work. Then, we introduce our approach of learning disentangled representations via human feedback and provide some theoretical guarantees. Before concluding, we present the results of our empirical evaluation.

\section{Related Work}
Our work touches upon correcting machine learning models and learning disentangled representation. 

\textbf{Correcting Machine Learning Models.} Machine learning models trained on biased data may exhibit undesirable bias due to data-collecting process. A number of works have been done to correct machine learning models from learning this bias \cite{ross2017right, murdoch2018beyond, teso2019explanatory, rieger2019interpretations, erion2019learning, selvaraju2019taking, schramowski2020making, shao2021right}. However, these works can only remove very simple bias on the observational feature level. What if the bias is on a more abstract level? \citet{kim2019learning} propose to unlearn more complex bias in a data set by minimizing the mutual information between the transformed feature and the target bias. These works are all restricted to discriminative models.

\textbf{Disentangled Representation.} In generative modelling, learning disentangled representations is an active area of research that receives increasingly attention \cite{higgins2016beta, kumar2017variational,  kim2018disentangling, chen2018isolating, burgess2018understanding, kim2018disentangling, mathieu2019disentangling}. $\beta$-VAE \cite{higgins2016beta} is one of the earliest work of this kind, which augmented the lower bound formulation to regulates the strength of independence prior pressures. FactorVAE \cite{kim2018disentangling} encourages the distribution of representations to be factorial and hence independent across the dimensions. $\beta$-TCVAE \cite{chen2018isolating} proposes an equivalent objective as FactorVAE but optimize it differently. However, these unsupervised learning of disentangled representations is shown to be fundamentally impossible from i.i.d.\ observations without inductive biases or some form of supervision \cite{locatello2019challenging}.  Thereafter this research problem starts to be addressed with supervision to provide general solutions \cite{ridgeway2018learning, bouchacourt2017multi, hosoya2019group, chen2019weakly,locatello2019disentangling, shu2020weakly, sorrenson2020disentanglement, khemakhem2020variational}. \citet{chen2019weakly} use weak supervision by providing similarities between instances based on a factor to be disentangled and formulate a regularized ELBO to enforce disentanglement. \citet{shu2020weakly} propose a theoretical framework to analyze the disentanglement guarantees of weak supervision algorithms. In particular, they decompose disentanglement into consistency and restrictiveness. \citet{locatello2020weakly} also use paired observations as weak supervision and they require weaker assumptions than \cite{shu2020weakly}. \citet{locatello2019disentangling} use a small number of labels to learn disentangled representations.

\begin{figure*}[t]
    \centering
    \includegraphics[scale=0.55]{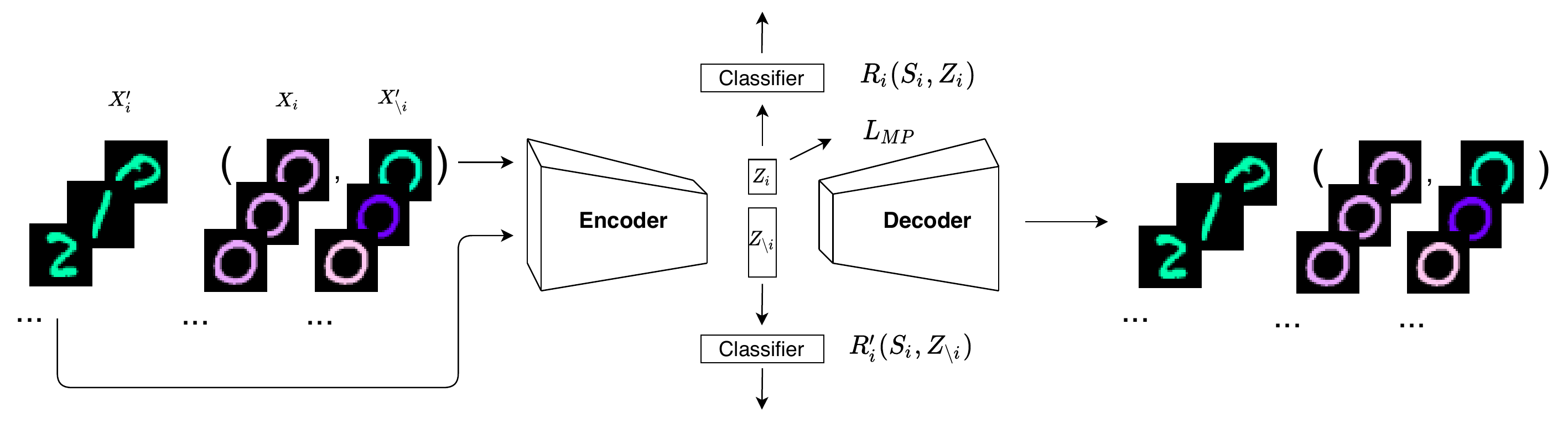}
    \caption{Our VAE uses a small amount of human feedback in addition to the train set to learn disentangled representations. For the sake of simplicity, only augmented loss terms are illustrated here. } 
    \label{fig:vae_structure}
\end{figure*}

\section{Unbiased Generative Models via Disentanglement}
\label{sec:unbiasedGenerativeModelviaDisentanglement}

We formalize the debiasing setting in the following way.
Let $S_1, \dots S_n$ denote independent latent factors underlying the variation of the data, and let them
follow some prior distribution $p(S) = \prod_i p(S_i)$.
We assume that they produce the observations $X$ via a deterministic function $g$, i.e., $X= g(S)$, yielding
a distribution $p(X)$, and that $S$ can be recovered via an encoder
$S = e(X)$.
Unfortunately, our training set suffers from some bias, giving rise to unwanted correlations between
the factors of variation as expressed by some other distribution $p'(S)$.
We merely assume that it is faithful to the true distribution on the level of single variable marginals,
i.e., $p'(S_i) = p(S_i)$, which is typically relatively easy to ensure.
We call the resulting biased distribution of our training data $p'(X)$.

Our goal is to obtain a generative model of $p(X)$ based on samples from $p'(X)$ in combination with small number
of examples providing weak supervision. We follow the VAE framework, introducing a latent code
$Z = Z_1, \dots, Z_m$ with $m \geq n$, and fix an independent prior distribution over the $Z_i$, e.g., a unit Gaussian.
Via a learned decoder $\hat{g}(Z)$, we would like to generate samples matching $p(X)$, i.e.
$p(\hat{g}(Z)) \approx p(X)$.

Suppose that we have managed to perfectly disentangle the VAE's latent codes 
with regard to the true factors of variation, by ensuring that each $Z_i$ captures exactly $S_i$ for $i \leq n$.
Then there exist bijections $f_i$ deterministically mapping the two to each other: $Z_i = f_i(S_i)$,
and their marginal probabilities will match, i.e., $p(S_i) = p'(S_i) = p(f(S_i)) | \frac{d}{d S_i} f(S_i) | $.

If in addition, the VAE's decoder $\hat{g}$ approximately matches the data generating procedure such that
$\hat{g}(f(S)) \approx g(S)$,
then it follows that the distribution $\hat{p}(X)$ represented by the VAE matches the desired unbiased distribution
$p(X)$. Letting $\delta_X[\cdot]$ denote the Dirac delta at $X$, we have
\DeclarePairedDelimiter\abs{\lvert}{\rvert}
\begin{align}
    p(X) &= \int \delta_X[g(S)] \prod p(S_i) dS \displaybreak[3]\\
    &\approx \int \delta_X[\hat{g}(f(S))]\prod p(f(S_i))  \abs*{\frac{d}{d S_i} f(S_i)} dS \\
    &= \int \delta_X[\hat{g}(Z)] \prod p(Z_i) dZ = \hat{p}(X),
\end{align}
where the last line follows from substituting $Z = f(S)$, cancelling out the differentials.

\begin{figure}[t]
	\centering
	\begin{tikzpicture}[latent/.append style={minimum size=1.0cm}]

	\node[ latent]                    (snoti) {$S_{\setminus i}$};
	\node[ latent, right= of snoti]       (si) {$S_i$};
	\node[ latent, right= of si]       (zi) {$Z_i$};
	\node[ latent, right= of zi]       (znoti) {$Z_{\setminus i}$};

	\node[ obs, below=0.6 cm of si, xshift=-0.9cm] (x1) {$X$};
	\node[ obs, below=0.6 cm of znoti, xshift=-0.9cm] (x2) {$X$};
	\edge{si, snoti}{x1};
	\edge{zi, znoti}{x2};
	
	\draw[dashed, ->] (zi) -- node[below] {$R_i$} (si);
	\draw[dashed, ->] (znoti) edge[bend right]  [strike through] node[above, pos=0.35]  {$R_i'$} (si);
	\draw[dashed, ->] (snoti) edge[bend left] [strike through] node[above, pos=0.35]  {$L^i_{MP}$}  (zi);
	\end{tikzpicture}
	\caption{Illustration of the purpose of the disentanglement losses. The match-pairing loss prevents
	unrelated factors of variation $S_{\setminus i}$ from having an influence on $Z_i$. In turn, the positive
	classification loss $R_i$ ensures that $Z_i$ is predictive for $S_i$, while the negative
	classification loss $R_i'$ ensures that $Z_{\setminus i}$ is not.}
	\label{fig:model}
\end{figure}
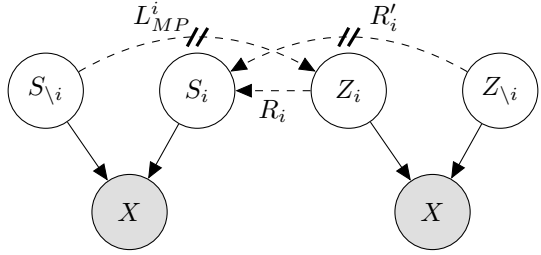

\subsection{Disentanglement through Human Feedback} 
Based on this insight, we design a combination of loss functions to encourage the model to fit the training
data as well as possible while enforcing disentanglement. The former is achieved using the standard ELBO loss.
To ensure that a latent variable $Z_i$ only contains information pertaining to target factor $S_i$, we collect
pairs of samples $X, X'$ which share the same value for $S_i$, while the other factors $S_{\setminus i}$
vary at random. Following \citet{shu2020weakly}, we minimize the match-pairing loss
$L^i_{MP}(X, X') = \mathbb{E}[(Z_i - Z_i')^2]$ for those samples, where
$Z_i, Z_i'$ are sampled from $q(Z_{i}|X), q(Z_{i}|X')$. In addition, we use a small number of samples $X$ for which the true value for the target underlying factor $S_i$ is available.
To ensure that $Z_i$ is predictive for $S_i$, while the rest of the latent variables $Z_{\setminus i}$ is not,
we train regression models in parallel with the main model to predict $S_i$ from
$Z_i$ and $Z_{\setminus i}$ respectively.
We then use these models to compute the classification loss
$L^i_{CL}(X) = \mathbb{E}_{q(Z|X)}[\lambda_2 R_{i}(S_i, Z_{i}) - \lambda_3 R_{i}'(S_i,  Z_{\setminus i})]$,
where $R_{i}$ and $R'_{i}$ denote the cross-entropy classification loss of the two classifiers.
The way these losses interlock is illustrated in Fig. \ref{fig:model}. The full loss for our disentangled VAE is $ L(X) =$
\begin{align}
    -\text{ELBO}(X) + \sum\nolimits_i \lambda_1 L^i_{MP}(X) + L^i_{CL}(X),
    \label{eq:loss}
\end{align}
where $L^i_{MP}$ and $L^i_{CL}$ is only computed for all datapoints and facors of variation $i$ for which the required augmentations are available.
Fig. \ref{fig:vae_structure} illustrates the resulting computation graph.

\subsection{Analysis of Theoretical Guarantees}
Following the arguments from \citet{locatello2019challenging}, we prove that disentanglement arises from our supervision method instead of model inductive bias.
We use the definition of disentanglement from \citet{shu2020weakly} where
disentanglement is decomposed into \emph{consistency} and \emph{restrictiveness}.
Strong disentanglement is guaranteed when both consistency and restrictiveness hold. Let $\hat{e}_i$ denote a learned deterministic encoder mapping from $X$ to
$Z_i$.
According to the definition of \citet{shu2020weakly}, $S_{i}$ is \emph{consistent} with $Z_{i}$ if $\mathbb{E} [||\hat{e}_{i} \circ g(S_{i}, S_{\setminus i}) - \hat{e}_{i} \circ g(S_{i}, S_{\setminus i}')||^2] = 0 $, and $S_{i}$ is \emph{restricted} to $Z_{i}$ if $\mathbb{E}\left[ ||\hat{e}_{\setminus i} \circ g(S_{i}, S_{\setminus i}) - \hat{e}_{\setminus i} \circ g(S_{i}', S_{\setminus i})||^2\right]= 0 $. Intuitively, consistency expresses that $Z_{i}$ does not capture the factors of variation other than $S_{i}$, and restrictiveness expresses that $S_{i}$ is only captured in $Z_{i}$. 
In addition, \citet{shu2020weakly} discuss the property that
$Z$ should be \emph{non-trivial}, i.e., it should encode $S$ at all.

\begin{table*}[tb]
\caption{Properties of our and match pairing approach in case of disentangling a subset of factors (shortly denoted as ``subset'' in the Tab.) and disentangling all factors (shortly denoted as ``all factors'' in the Tab.)}
\vskip 0.1in
\centering
\begin{sc}
\resizebox{.9\textwidth}{!}{
\begin{tabular}{lccccc}
\toprule
 & & consistency & restrictiveness & \thead{disentanglement \\ (consistency $\bigwedge$ restrictiveness)} & non-trivial guarantee \\
 \midrule
 \multirow{2}{*}{Ours} & subset & \ding{51} & \ding{51} &\ding{51} & \ding{51}\\ 
 & all factors & \ding{51} & \ding{51} & \ding{51} & \ding{51} \\
 \midrule
 \multirow{2}{*}{\thead{Match \\ Pairing}} & subset & \ding{51} & \ding{55} & \ding{55} & \ding{55}\\ 
 & all factors & \ding{51} & \ding{51} & \ding{51} & \ding{51} \\
 \bottomrule
\end{tabular} 
}
\end{sc}
\label{tab:summary_property}
\end{table*}

Tab.~\ref{tab:summary_property} summarizes the theoretical guarantees of each property of our method and pure match pairing. We distinguish two cases: Disentangling \emph{all} factors, and
disentangling only a subset of factors, leaving the rest as
arbitrarily encoded nuisance factors. Since our method builds on match pairing, its guarantees trivially
transfer. Therefore, we only need to prove that the additional
label-based supervision we introduce is sufficient for learning
restrictive and non-trivial solutions when disentangling a subset of factors.

Define a hypothesis space $\mathcal{H}$ of models we are willing to consider.
Let $(p(S), p'(S), g, e) \in \mathcal{H}$ denote an arbitrary ground-truth
model which generated the data we are observing.
Following \citet{shu2020weakly}, we call a supervision method 
$S : \mathcal{H} \to \mathcal{P}$
\emph{sufficient}
for a certain guarantee if there exists a learning algorithm
$\mathcal{A} : \mathcal{P} \to \mathcal{H}$ which uses the supplied examples
to produce a learned model $(p(Z), \hat{g}, \hat{e})$ for which the desired
guarantee holds.
In the following, we show that our method of supervision is sufficient
for non-trivial and restrictive solutions.
We do so by considering an $\mathcal{A}^*$ which obtains
the global minimum of the classification loss $L^i_{CL}$, despite using
universal approximators for $R$ and $R'$.

\textbf{Non-Triviality Guarantee.} 
We claim that our method guarantees the learned model $(p(z), \hat{g}, \hat{e})$
provides non-trivial solutions for $Z_{i}$
, i.e. there is a learning algorithm that for which $S_i \not\!\perp\!\!\!\perp Z_i$.
Match pairing alone does not entail this guarantee when disentangling only a
subset of factors. To see why, consider a simple counterexample with underlying
factors $S_{1}, S_2 \sim \mathcal{N}(0, 1)$, and generator $g(S) = S_1 \cdot S_2$.
There exists a learning algorithm
$\mathcal{A}$ which optimizes the match pairing loss on factor
$S_{1}$ by setting
$(g(S_{1} , S_{2}), g(S_{1} , S^{'}_{2}))  = (\hat{g}(Z_{1} , Z_{2}), \hat{g}(Z_{1} , Z^{'}_{2}))$,
and $\hat{g}(Z)=[Z_{1}, Z_{2}] = [0, Z'_{1} \times Z'_{2}]$
where $Z'_{1} \sim \mathcal{N}(0, 1)$, $Z'_{2} \sim \mathcal{N}(0, 1)$. In this case, match pairing yields a trivial solution for $Z_{1}$, i.e. $Z_{1} = 0$.

\emph{Proof}: A learned model $(p(z), \hat{g}, \hat{e})$ provided by
$\mathcal{A}^*$ minimizes the classification loss, and the positive term
$R^i(Z_i, S_i)$ in particular. Consequently, it is possible to accurately predict
$S_i$ from $Z_i$, and $S_i \not\!\perp\!\!\!\perp Z_i$.
\qed

\textbf{Restrictiveness Guarantee.}
We postulate that our method guarantees that the learned model $(\hat{p}(z), \hat{g}, \hat{e})$ restricts the information of $S_{i}$ to $Z_{i}$. That is, $Z_{\setminus i} \perp \!\!\!\perp S_{i}$.

\emph{Proof}: Suppose to the contrary that $Z_{\setminus i}$ contains information about $S_{i}$.
Then we could find a regression model that predicts $S_i$ better than random,
i.e., the value of $-R'(S_i, Z_{\setminus i})$ would be suboptimal.
This is a contradiction, since we assumed that $\mathcal{A}^*$
would achieve the global minimum of $L^i_{CL}$. \qed

\begin{table}[t]
\caption{Summary of the datasets. Parentheses give number of quantized values for each target factor. The last column gives the size of samples we chose from the train set to perform data augmentation on. }
\vskip 0.1in
\centering
\begin{sc}
\resizebox{.8\columnwidth}{!}{
\begin{tabular}{lccccc}
\toprule
 Name & \thead{Training \\ instances} & \thead{Held-out \\ instances} & Image size & \thead{ground-truth \\ factor} & \thead{Data \\ Augmentation}\\
 \midrule
\thead{colored \\ MNIST} & 60k & 10k & 28 $\times$ 28 $\times$ 3 & \thead{shape(10), \\ color(10)} & 600 \\
\thead{colored \\ dSprites} & 10k & 10k & 64 $\times$ 64 $\times$ 3 & \thead{shape(3), color(3), \\ nuisance} & 1k\\
3d shapes & 10k & 10k & 64 $\times$ 64 $\times$ 3 & \thead{object shape(4), \\ object color(4), \\ nuisance} & 1k\\
 \bottomrule
\end{tabular} 
}
\end{sc}
\label{tab:data_set_summary}
\end{table}

\section{Empirical Investigation}
Our main intention here is to address the following question empirically: 
{\it Does our proposed model disentangle the latent space better than the mainstream counterpart models? } To this end, we ran a series of experiments for both qualitative and quantitative evaluation. For the qualitative comparison, we considered reconstructed images and latent space arithmetic for generating novel content, as well as latent space traversal.

\textbf{Dataset.} 
We considered several standard benchmark datasets. First, we generated colored MNIST \cite{lecun2010mnist} in a similar way as \cite{kim2019learning}. In this domain, we assumed two latent factors of variations: shape and color. In train set, every shape class of images was randomly and consistently assigned with a color so the shape class and the color class correlated with each other. In test set, one could assign random colors to each image. But we made the problem even more difficult by assigning the color class reversely correlated to the shape class so that none of these combinations of shape and color were seen in the train set. As another dataset, we generated colored dSprites \cite{higgins2016beta} in a similar spirit. We randomly sampled 10k instances as train set and injected confounding in it by consistently assigning one color to each shape category and the rest factors remained untouched. In test set, we assigned each color to the shape with one offset as to in the train set, so that none of these two factors' combinations were seen in the train set. The third dataset is 3D Shapes \cite{kim2018disentangling}. We sampled 10k instances to be the train set whereby each object hue category correlated to one object shape category. In test set, each object hue category correlated to the object shape category with one offset as to in the train set for the same reason.

Tab.~\ref{tab:data_set_summary} summarizes the datasets, and
Fig.~\ref{fig:overview} (left) shows a whole spectrum of the colored MNIST domain based on one random run (colors are randomized for each run on colored MNIST and colored dSprites). The category of the target latent factors for the train set are highlighted in orange rectangles.

\begin{figure*}
    \centering
    \includegraphics[scale=0.44]{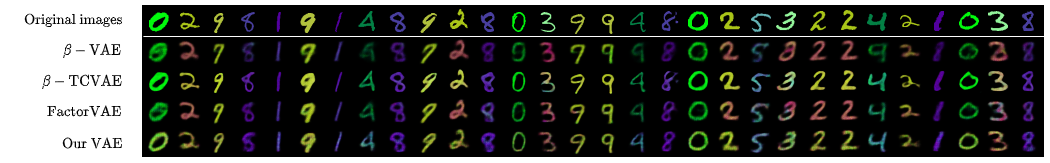} 
    
    \includegraphics[scale=0.44]{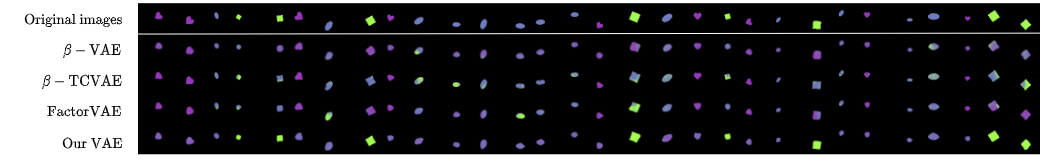}
        
    \includegraphics[scale=0.44]{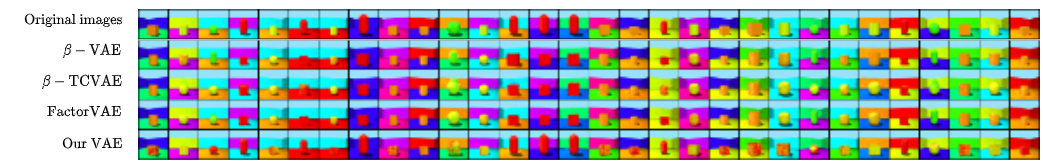}
    
    \caption{Reconstruction images from different models on colored MNIST, colored dSprites and 3D shapes. Each shape factor is consistently associated with one color factor in the train set, while in the test set the association is different. Some images from the baseline models are reconstructed wrongly with the shape (color) factor that associated with the corresponding color (shape) factor in the train set, implying entangled latent representations.}
    \label{fig:reconstruction_mnist}
\end{figure*}

\begin{figure*}[t]
    \centering
    \begin{minipage}{0.69\textwidth}
    \centering
    \includegraphics[width=0.95\textwidth]{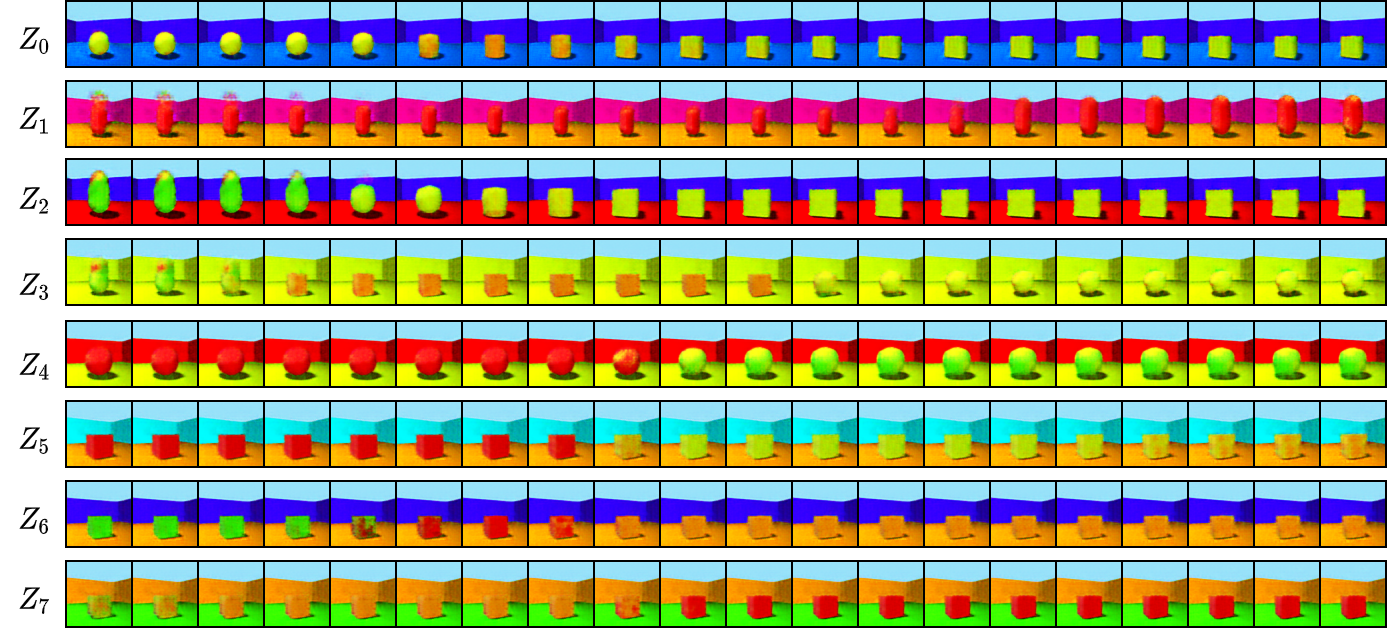}
    \end{minipage}
    \caption{Latent space traversal of our VAE. The first (last) 4 rows are respectively the traversal of $Z_{0}$, $Z_{1}$, $Z_{2}$ and $Z_{3}$ ($Z_{4}$, $Z_{5}$, $Z_{6}$ and $Z_{7}$) which are collectively intended for shape (color) factor. The samples show that varying the first (last) 4 variables indeed leads to varying object shape (color), which suggests that each unit of representation has captured the intended factor. Meanwhile, the nuisance factors such as wall color stay unchanged, which suggests that the nuisance factors are not captured in these 8 variables.  \label{fig:latent_traversal} } 
    
\end{figure*}

\textbf{Data Augmentation to Emulate a Human User.} We augmented the biased dataset with a few amount of examples to emulate human feedback. Specifically, on each target ground-truth factor $S_{i}$, we generate a few samples $X_{i}'$ that vary on the rest factors $S_{\setminus i}$ while keeping the target factor fixed, and a few samples $X_{\setminus i}'$ that vary only on the target factor while keeping the rest factors fixed. That is, $X_{i}'$ share the target factor and only the target factor, and $X_{\setminus i}'$ share the rest factors and only the rest factors. In addition, we acquired a few $S_{i}$ of $X_{i}'$ to avoid trivial solutions of $Z_{i}$ (non-trivial guarantee) by enforcing $Z_{i}$ to contain information about $S_{i}$, and we constrained that the nuisance variables $Z_{\setminus i}$ do not contain information about $S_{i}$ (restrictiveness guarantee). 

\begin{figure*}[t]
    \centering
    \includegraphics[scale=0.25]{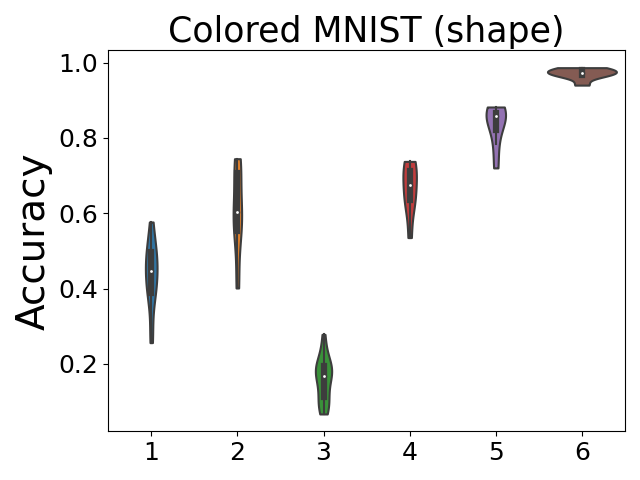}
    \includegraphics[scale=0.25]{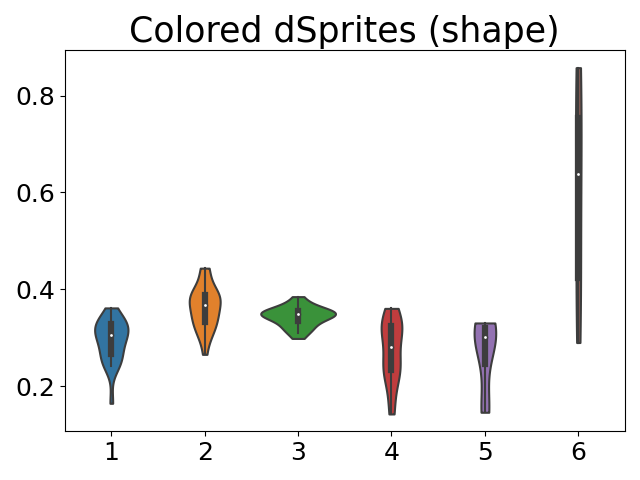}
    \includegraphics[scale=0.25]{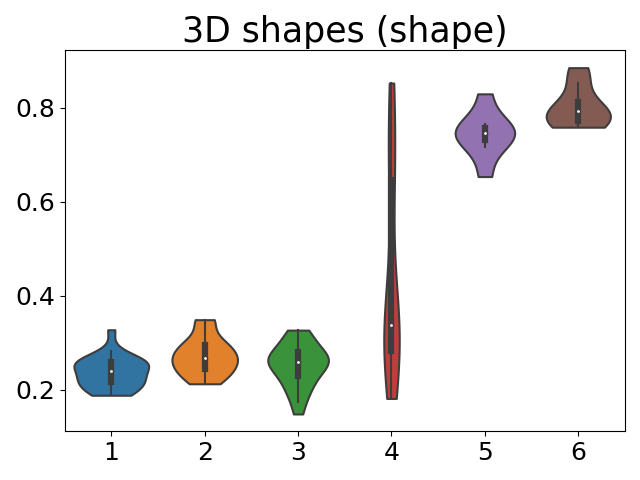} \\
    \includegraphics[scale=0.25]{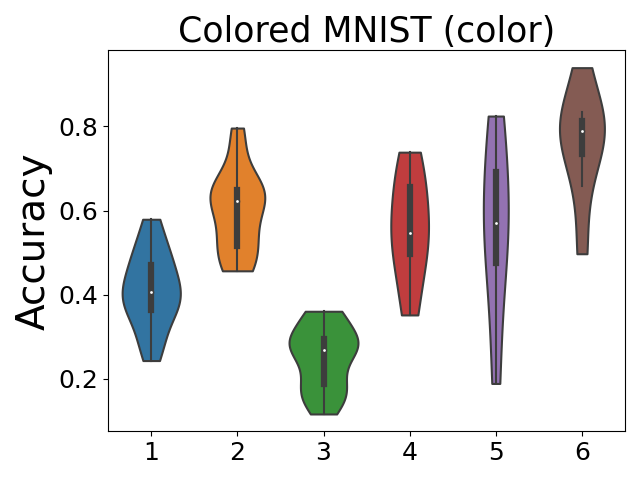}
    \includegraphics[scale=0.25]{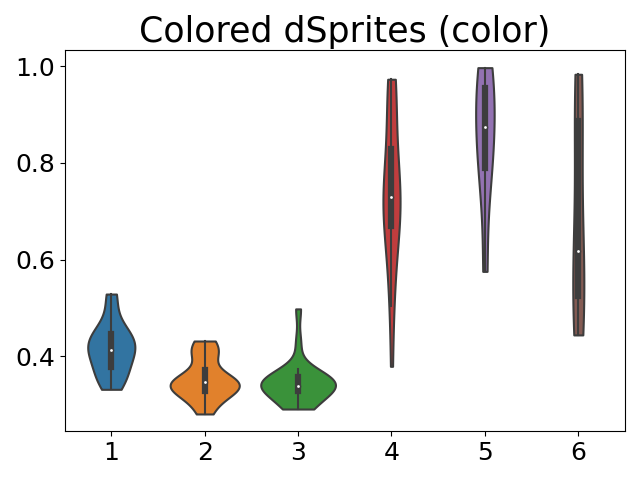}
    \includegraphics[scale=0.25]{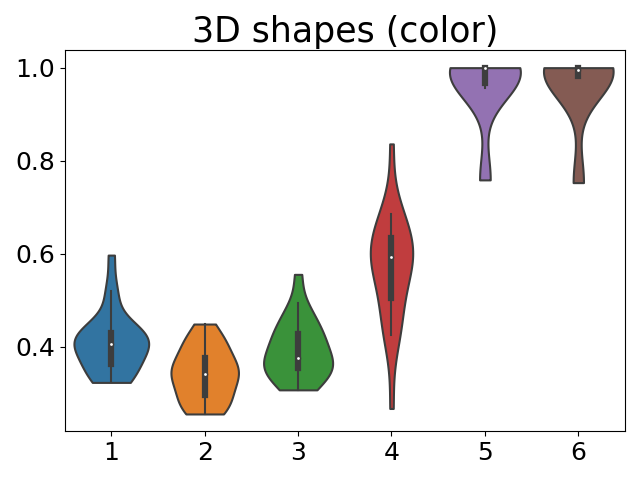}
    \caption{Accuracies of the post-hoc linear classifiers after debiasing for shape (top row) and color (bottom row).  1: $\beta$-VAE, 2: $\beta$-TCVAE, 3: FactorVAE, 4: Ada-GVAE, 5: Our VAE without annotations, 6: Our VAE with annotations. The higher the value, the better the performance. }
    \label{fig:accuracy_downstream}
\end{figure*}

\begin{figure*}[t]
    \centering
    \includegraphics[scale=0.25]{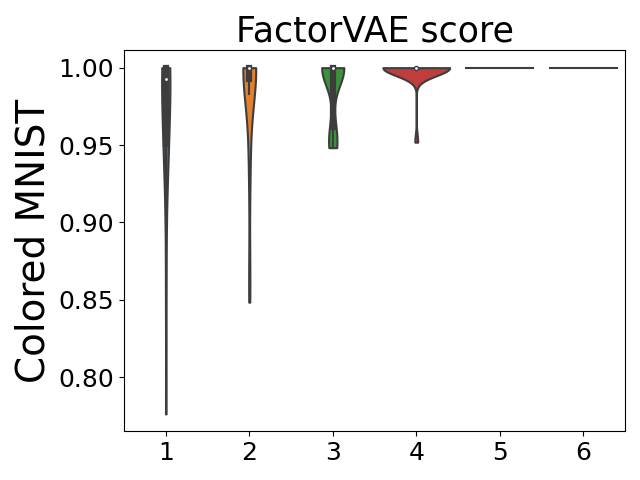}
    \includegraphics[scale=0.25]{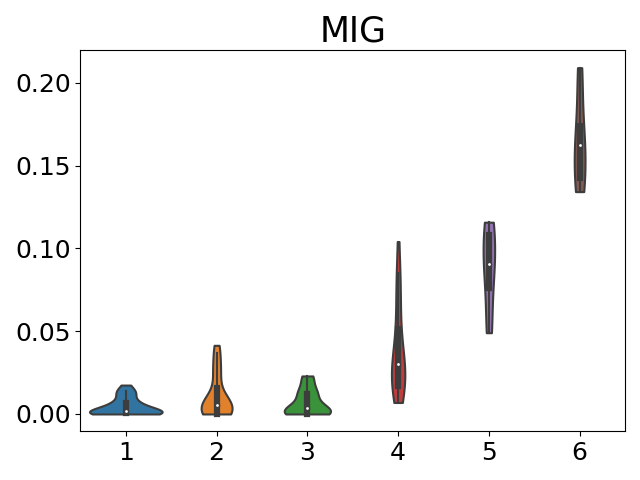}
    \includegraphics[scale=0.25]{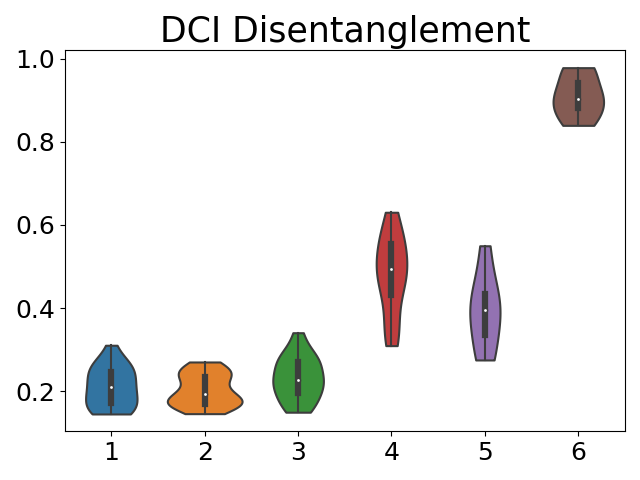}
    \includegraphics[scale=0.25]{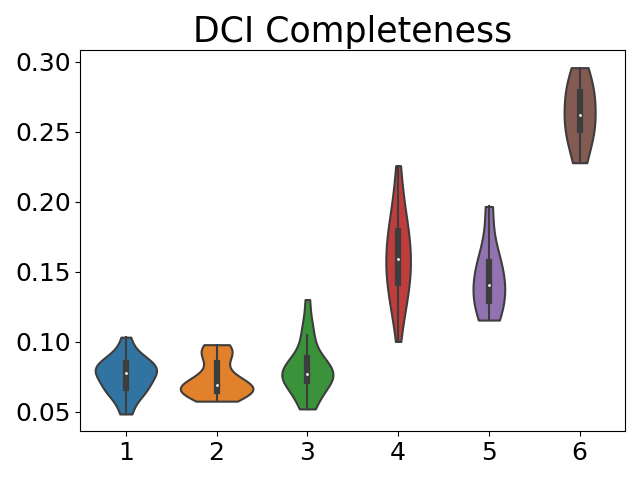}
    \includegraphics[scale=0.25, trim={0cm, 0cm, 0cm, 1.25cm}, clip]{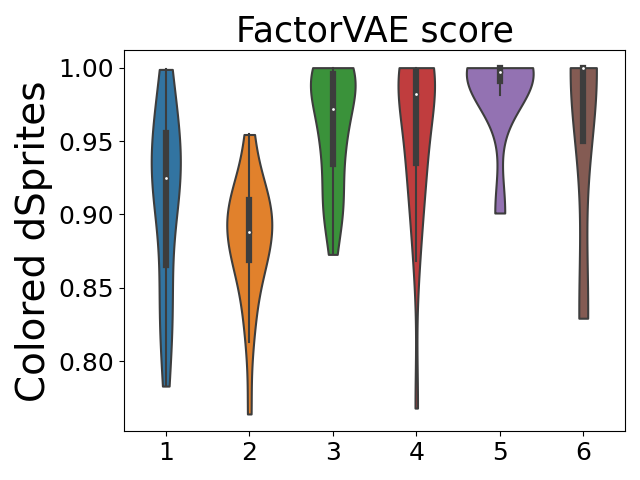}
    \includegraphics[scale=0.25, trim={0cm, 0cm, 0cm, 1.25cm}, clip]{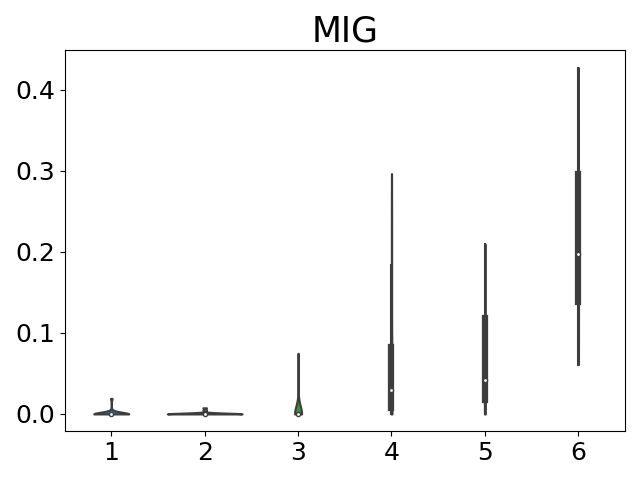}
    \includegraphics[scale=0.25, trim={0cm, 0cm, 0cm, 1.25cm}, clip]{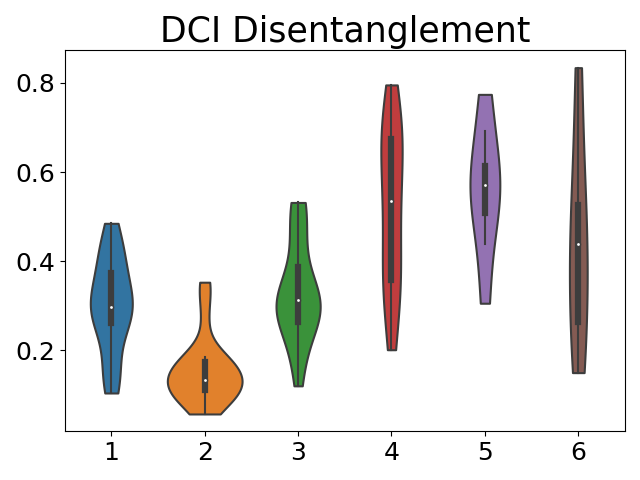}
    \includegraphics[scale=0.25, trim={0cm, 0cm, 0cm, 1.25cm}, clip]{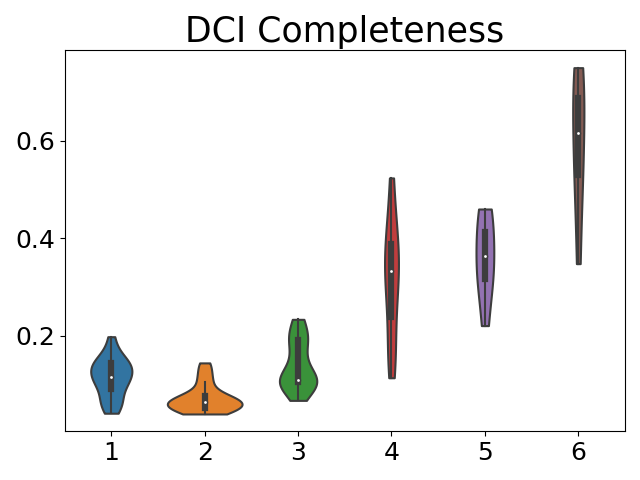}
    
    \includegraphics[scale=0.25, trim={0cm, 0cm, 0cm, 1.25cm}, clip]{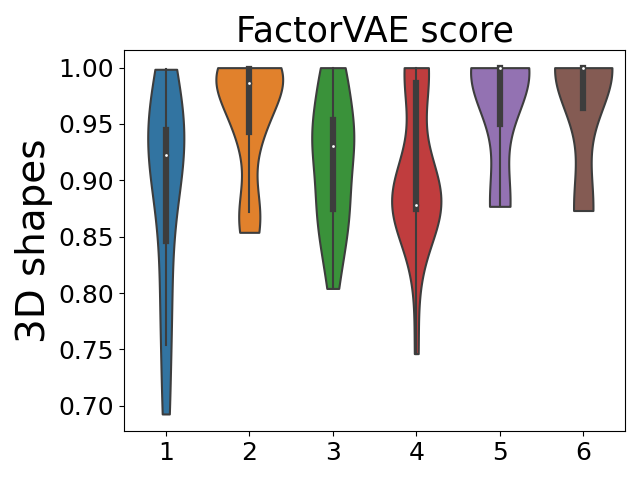}
    \includegraphics[scale=0.25, trim={0cm, 0cm, 0cm, 1.25cm}, clip]{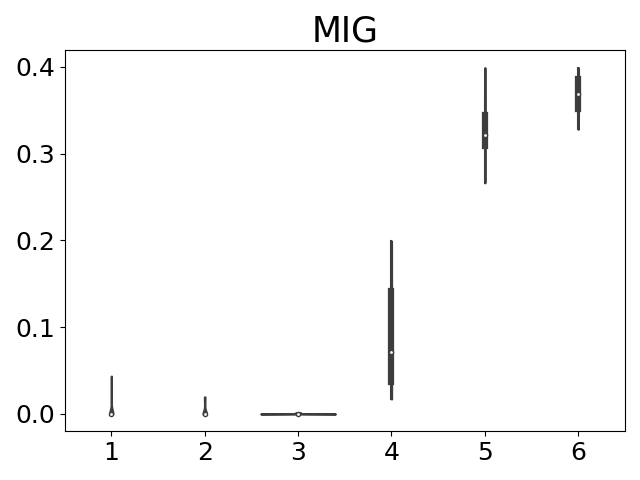}
    \includegraphics[scale=0.25, trim={0cm, 0cm, 0cm, 1.25cm}, clip]{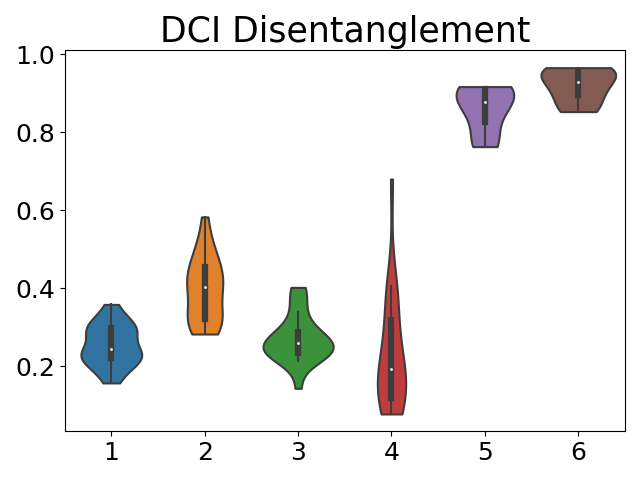}
    \includegraphics[scale=0.25, trim={0cm, 0cm, 0cm, 1.25cm}, clip]{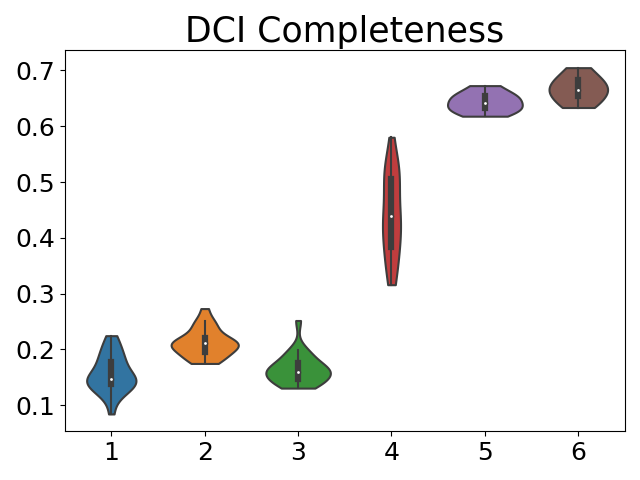}
    \caption{Several disentanglement metrics (FactorVAE score, MIG, DCI disentanglement and DCI completeness) across different models on colored MNIST (top row), colored dSprites (middle row) and 3D shapes (bottom row). 1: $\beta$-VAE, 2: $\beta$-TCVAE, 3: FactorVAE, 4: Ada-GVAE, 5: Our VAE without annotations, 6: Our VAE with annotations. The higher the value, the better the performance. }
    \label{fig:disentanglement_metrics}
\end{figure*}

\textbf{Experimental Protocol.} We followed the experimental protocol of  \cite{locatello2019disentangling, locatello2020weakly} and aligned with the general literatures on disentanglement \cite{higgins2016beta, kim2018disentangling, chen2018isolating}. The baseline models we used are $\beta$-VAE \cite{higgins2016beta}, FactorVAE \cite{kim2018disentangling}, $\beta$-TCVAE \cite{chen2018isolating} and Ada-GVAE \cite{locatello2020weakly}. These models are mainstream VAEs for benchmarking disentangled representations. In addition, we also evaluated our VAE with unannotated human feedback to investigate the value of the annotations. We implemented all models in python and pytorch. For each model, we ran experiments with 8 random seeds. For each baseline model, we used 3 fixed hyperparameters on each dataset. Our model's hyperparameters were chosen empirically according to the quality of the reconstructions. Our experience suggests using hyperparameters $\lambda_{1} = \lambda_{2}$ and $\lambda_{3} = 1$ or $\lambda_{3} = 10$ lead to good results in practice, so we propose to tune the hyperpamraters within this range. We used linear classifiers for $R_{s}$ and $R_{s}'$ in \eqref{eq:loss}. We relaxed the condition for $R'_{s}$ to be an universal classifier in practice and demonstrated that this can still yield promising disentangled representations empirically.

\subsection{Qualitative Evaluation}
To gain a qualitative and intuitive insights, let us start off by investigating reconstructed images. 

{\bf Reconstructions.}
As mentioned above, we chose the test samples to be completely new to the trained models by sampling on the latent space where the combinations of the target latent factors were never seen in the train set. In this case, if the target latent factors were entangled in the learned representations, the model would struggle to encode all the target latent factors in this unseen combination, this in turn would lead to unsatisfactory reconstructions. Fig. \ref{fig:reconstruction_mnist} shows some randomly chosen test samples across the three datasets and their reconstructions via different models. One can see for colored MNIST and colored dSprites, the reconstructed color via the baseline models for some samples still exhibited the color that associated with those samples' shape in the train set. However, this happens much less often on our model. On 3D shapes, the object color is mostly reconstructed reasonably, but for the baseline models, the reconstructed object shapes often still exhibit the shape that associated with the corresponding object color in the train set, implying for entangled latent representations. For our model, both the object shape and the object color were reconstructed very plausibly. The model achieved this robustness w.r.t.~generalization on the unseen samples by disentangling the latent representations.

\begin{figure}[t]
    \centering
    \includegraphics[width=0.2\columnwidth]{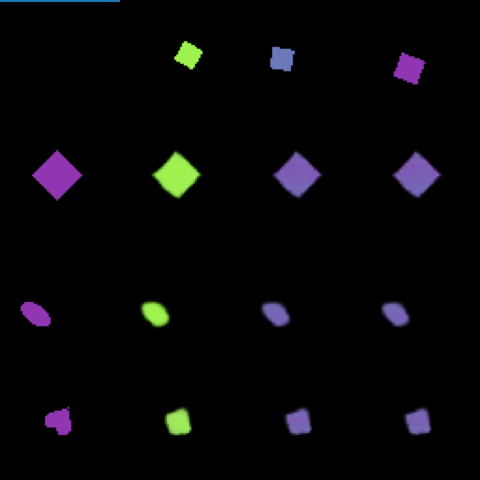} 
    \includegraphics[width=0.2\columnwidth]{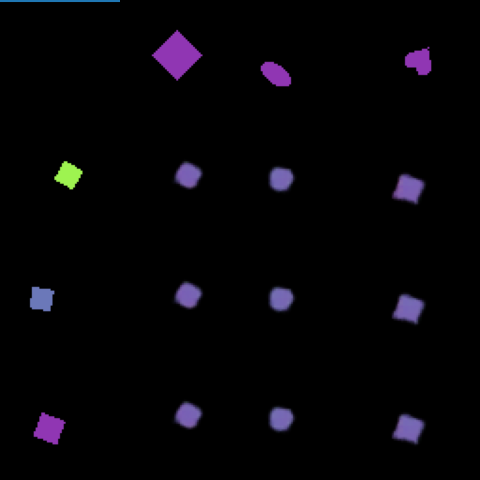}
    
    \vspace{0.05cm}
    \hspace{0.01cm}
    \includegraphics[width=0.2\columnwidth]{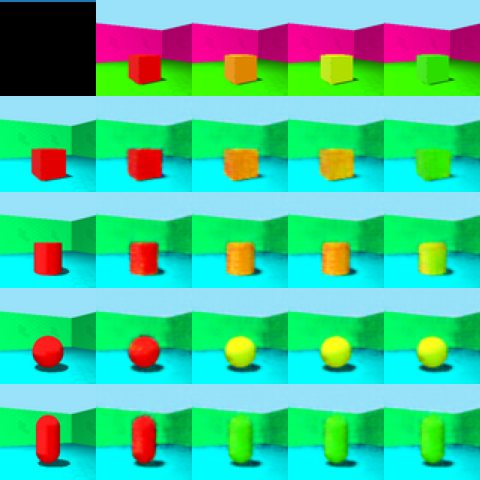}
    \includegraphics[width=0.2\columnwidth]{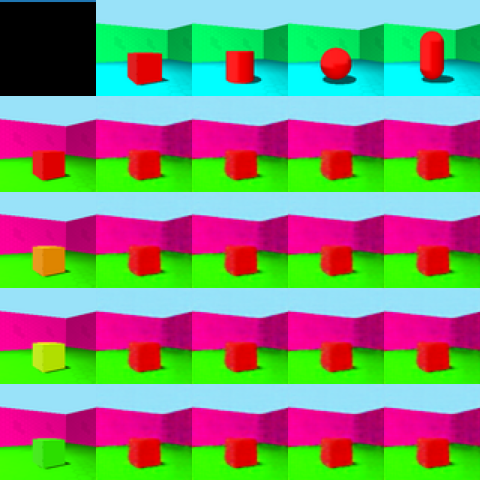}
    \caption{Cross-product of shapes (leftmost column) and colors (top row) generated by the debiased VAE.}
    \label{fig:latent_arithmetic}
\end{figure}

{\bf Latent Space Arithmetic.}
This experiment demonstrates that our model can not only generalize under non-i.i.d distributions, but can also generate novel content. Figs. \ref{fig:overview} (right) and \ref{fig:latent_arithmetic} show all possible hybridization of shapes and colors generated by our VAE in each corresponding domain. Each hybridized image was generated by concatenating the latent representation of one image's shape on the leftmost column and another image's color on the first row. When nuisance variables are present, they take the values of the average of both reference images' nuisance representations. The result shows that the hybridized images very plausibly resemble both the shape from the leftmost-column images and the color from the first-row images, and this yields and recovers the whole spectrum of the target domain where only a small fraction of which were seen in the train set. This not only allows us to generate novel contents, but also grants us opportunities to manipulate the latent space and generate specific samples for downstream tasks.

{\bf Latent Space Traversal.}
To give a better overview of the latent space and what the generative models have really learnt for the target factors, we show partial latent space traversals of our model for 3D shapes. The latent variables are in total 50. We designated the first 4 variables for learning the shape factor, and the second 4 variables for the color factor, the rest are considered nuisance variables. Fig. \ref{fig:latent_traversal} shows samples generated by traversing the first 8 latent variables. The first (last) 4 rows correspond to the variables that are supposed to learn the shape (color) factor. The samples suggest that each unit of the representations has reasonably learnt the target factor and is also mostly restricted to that factor. Moreover, the nuisance factors, wall color etc, seem not to be captured by these 8 variables which implies the latent representations are consistent with the target factors.

\subsection{Quantitative Evaluation}
For a quantitative evaluation of the disentanglement, we used FactorVAE score \cite{kim2018disentangling}, DCI Disentanglement and DCI Completeness \cite{eastwood2018framework} and Mutual Information Gap (MIG) \cite{chen2018isolating}. As these metrics are not conceptualized accounting for biased data, we use the whole spectrum of the target domain to evaluate them.

DCI Disentanglement measures the degree to which each latent variable capturing at most one generative factor. This score gives a direct hint on the consistency property (see Tab. \ref{tab:summary_property}). DCI Completeness measures the degree to which each factor is captured by a single latent variable. Although our model does not assume single latent variable for each factor and therefore this metric is not in favor of us, this score still gives a hint on how densely the information of $S_{i}$ is packed in the latent variables, and in turn on the restrictiveness property (see Tab. \ref{tab:summary_property}). 

Since MIG enforces axis-alignment and our VAE does not make this assumption, this metric in its original form is not fit for evaluating our VAE. Therefore we adapt MIG by constraining the gap computation between the known unit of the latent representations and the rest representations. In particular, we slightly reformulate it to the following form: 
\begin{equation*}
\dfrac{1}{n} \sum\nolimits_{i=1}^{n} \dfrac{1}{H(S_{i})}(\max I(Z_{i}; S_{i}) - \max I(Z_{\setminus i}; S_{i}))
\end{equation*}
where $I(Z_{i}; S_{i})$ denotes the mutual information between the ground-truth factor $S_{i}$ and the unit of the latent representations designated for $S_{i}$, i.e.\ $Z_{i}$. $I(Z_{\setminus i}; S_{i})$ denotes the mutual information between $S_{i}$ and the rest of the latent representations $Z_{\setminus i}$. The higher this adapted MIG is, the information about $S_{i}$ is more densely packed in $Z_{i}$. Since the baseline models make the axis-alignment assumption, we use the original form of MIG to evaluate them. 

Fig. \ref{fig:disentanglement_metrics} shows the distribution of these scores from different models across different datasets via violin graphs. All the metrics are bounded by 0 and 1. Higher values imply better performance. As one can see, our method yields significantly better performance in terms of all the metrics. In terms of FactorVAE score, although the baseline models have also achieved almost the upper bound of FactorVAE score by its best performance, our model still has less variance across different runs, especially on coloresd MNIST we reached 1 for every random run. This demonstrates that our method is significantly better at learning disentangled representation on the considered datasets.

{\bf Downstream Accuracy.}
For each VAE, we trained a post-hoc linear classifier on each unit of the latent representations to predict the values of the corresponding factor. We use the original train set (without augmented data) where each shape is consistently associated with one color so that we can fairly compare the quality of the learned representations and exclude the influence of differently distributed train data for each classifier. For evaluation, we generated data where shape and color have different associations. This way, we make the test set a different distribution to the train set on purpose, so we can compare whose representations allow better generalization and robustness under covariate shifts \cite{quionero2009dataset}.

Fig.~\ref{fig:accuracy_downstream} reports the distribution of the test set accuracies. Apparently our model results in significantly better downstream accuracies, achieving 100\% accuracies at its best. This is a compelling evidence that our model yields more robust features for the downstream classifiers on the considered datasets.

\section{Conclusion and Future Work}
We presented a method for learning unbiased generative models using a biased training set which violates the i.i.d. assumption.
It obtains disentangled representations by incorporating additional human feedback in the form of a small amount of
examples with partially available label information. Our empirical study confirmed the effectiveness of our approach
at debiasing the data distribution, which in turn yields better generalization to unbiased test sets.

Our approach could benefit many real world tasks which should be investigated in future work.
One particular focus should be to make the learning process fully interactive.
In addition, one could work on generalizing this approach to other types of generative models.

\bibliography{ref}
\bibliographystyle{unsrtnat}

\end{document}